\pdfoutput=1

\documentclass[11pt]{article}

\usepackage[]{EMNLP2023}
\usepackage{times}
\usepackage{latexsym}
\usepackage{url}
\usepackage{hyperref}
\usepackage{nicematrix}
\usepackage{booktabs}
\usepackage{subfig}

\newcommand{\email}[1]{\ttfamily\begin{tabular}{@{} c @{}}#1\end{tabular}}

\usepackage[T1]{fontenc}

\usepackage[utf8]{inputenc}

\usepackage{microtype}

\usepackage{inconsolata}

\usepackage{graphicx}
\usepackage{float}
\usepackage{svg}
\usepackage{tabularx}

\usepackage{colortbl}
\usepackage{array}

\usepackage{listings} 
\usepackage{xcolor}   

\usepackage{todonotes}

\title{Balancing Transparency and Accuracy: A Comparative Analysis of Rule-Based and Deep Learning Models in Political Bias Classification}

\author{
\textbf{Manuel Nunez Martinez},
\textbf{Sonja Schmer-Galunder},
\textbf{Zoey Liu},\\
\textbf{Sangpil Youm},
\textbf{Chathuri Jayaweera},
\textbf{Bonnie J. Dorr}
\\
University of Florida, FL,USA \\
\email{\{manuel.nunez, s.schmergalunder, liu.ying, youms, \\ chathuri.jayawee, bonniejdorr\}@ufl.edu}
\\
}

\begin{document}
\maketitle
\vspace*{-.2in}
\begin{abstract}



The unchecked spread of digital information, 
combined with increasing political polarization and the tendency of individuals to isolate themselves from  opposing political viewpoints, has driven researchers to develop systems for automatically detecting political bias in media. This trend has been further fueled by discussions on social media. We explore methods for categorizing bias in US news articles, comparing rule-based and deep learning approaches. The study highlights the sensitivity of modern self-learning systems to unconstrained data ingestion, while reconsidering the strengths of traditional rule-based systems. Applying both models to left-leaning (CNN) and right-leaning (FOX) news articles, we assess their effectiveness on 
data beyond the original training and test sets.
This analysis highlights each model's accuracy,
offers a framework for exploring deep-learning explainability, and sheds light on political bias in US news media. We contrast the opaque architecture of a
deep learning model with the transparency of a
linguistically informed rule-based model, showing
that the rule-based model performs consistently across different data conditions and offers greater transparency,
whereas the deep learning model is dependent on the training set and struggles with unseen data.




\end{abstract}

\section{Introduction}
\vspace*{-.02in}

The current political climate in the United States is characterized by intense polarization and an unprecedented ease of publishing and disseminating information, where partisan hostility and negative perceptions of opposing party members are at an all-time high \cite{PEW-Research-article}. This dynamic is further exacerbated by social media platforms, where users curate their news feeds in a way that reinforces existing biases and isolates them from diverse perspectives, stifling constructive dialogue and creating what researchers term ``epistemic bubbles'' \cite{princeton-article}.

To address this, Natural Language Processing (NLP) researchers have developed models intended to automatically and objectively detect the presence and direction of bias.
Examples include model architectures ranging 
from rule-based designs
\cite{Hube2018} to State of the Art (SoA) transformer architectures
\cite{raza2024unlocking}.
While SoA architectures
have been shown to 
distinguish biased narratives from neutral ones, 
they struggle to learn the nuanced nature of bias expression without a sufficiently large and comprehensive dataset. 

\begin{figure}
    \centering
    \includegraphics[width=1\linewidth]{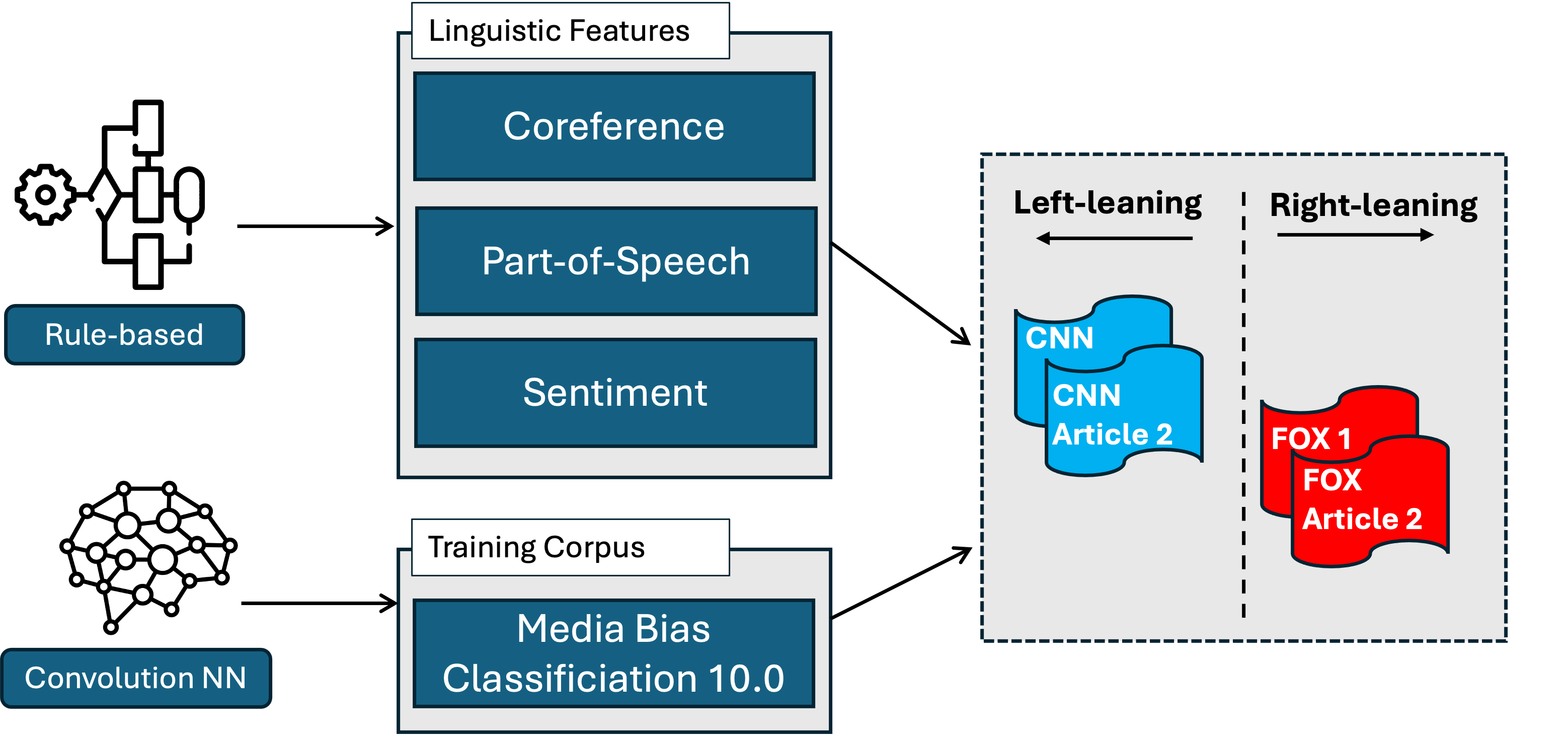}
    \caption{Comparison of Rule-based and Convolutional NN models: CNN and FOX news articles serve as external, unseen datasets for the Convolutional NN model. The rule-based model determines political bias using three linguistic features.}
    \label{fig:approaches}
    \vspace*{-.2in}
\end{figure}

Our contributions include an
investigation of both a rule-based and a deep learning model 
for 
political bias classification as depicted in Figure~\ref{fig:approaches}, 
with the goal of
promoting a more informed discussion
on bias detection methodologies.
To overcome data demands
of SoA architectures, we adopt a convolutional neural network model.\footnote{For brevity, we use ``convolutional NN model'' henceforth, as the abbreviation CNN is employed to refer to a news outlet.} Our contrasting approach
is a simpler, more transparent rule-based model for bias classification
using sentiment detection and linguistic features.
This model does not rely on
preexisting bias lexicons,
\textit{``black box''} machine learning models,
or
large 
training datasets. 
Moreover, its simplicity
allows 
for easy correction,
with a few, clearly delineated, components. 

A second contribution is the use of linguistic information for 
detecting an article's stance towards entities.\footnote{We define \textit{stance} as the overall attitude of a news article toward an entity, whereas \textit{sentiment} refers to a sentence-level (pos/neg) label.}
Our rule-based approach 
includes a novel part-of-speech driven ``reference resolution'' (e.g., associating adjectives with a corresponding noun), for a more focused stance assignment.
We emphasize that it is not our goal to achieve SoA performance for political bias classification through the rule-based model, but rather to explore the extent to which straightforward
linguistic features (parts of speech, coreference, and sentiment) can be leveraged to classify political bias. 

A third contribution involves
exploring methods to 
enhance
\textbf{explainability} 
of deep learning models. By testing a convolutional NN model on 
various datasets and correlating its performance disparities 
with differences in the data, we 
identify the features prioritized by
the model.

Our findings show that the rule-based model maintains consistent performance across various data conditions, presenting a clear right-leaning bias for FOX. By contrast, the convolutional NN model relies heavily on its training set, struggling with data not directly related to the political bias data on which it is trained. The rule-based approach performs comparably to deep learning in these situations, making it more applicable to real-world scenarios and offering greater transparency.




The next section reviews bias detection methodologies in news media. Section~\ref{sec:data} covers data collection, preprocessing, and experimental setup. Section~\ref{sec:model_implementation} details the implementation of rule-based and convolutional NN models. Section~\ref{sec:result} evaluates model performance and their application to external data, with concluding remarks in Section~\ref{sec:conclusion}.

\section{Related Work}
\label{sec:related_work}
Following \citet{Mullainathan2002MediaBM}, we view bias in news articles not as 
as a distortion or 
selective presentation of information 
to convey a belief, potentially impacting readers' opinions. 
%
Media bias
is categorized into coverage bias, gatekeeping bias, and statement bias \cite{Saez2013}. 
Our study focuses on
statement bias, 
i.e., the use of rhetoric describing entities
\cite{Hamborg2019}
identified by
our rule-based sentiment analysis model 
through 
identification of 
words conveying
sentiment toward entities. 

Entity Level Semantic Analysis (ELSA) \cite{ronningstad-etal-2022-entity}, 
is exemplified by
the work of \citet{Luo-2022}, 
where sentiment toward an entity is computed across sentences, iteratively lowering the sentiment scores for entities appearing in negative contexts.
Our current study adopts a form of ELSA that eliminates
the need for ``Negative Smoothing''
by using part-of-speech (POS) resolution 
to identify sentiment towards a given entity,
thus filtering
out ``noise'' 
introduced by incidental occurrences of nearby negative terms.
Deep-learning ELSA models
often suffer from an opaque architecture and 
overly broad feature selection.
\citet{fu-etal-2022-entity}
address this 
with a \textit{transparency layer} in
a convolutional NN, that adjusts
feature selection 
using an integrated gradient technique,
aligning with the POS resolution method described here.

Bias detection in media is typically handled
as binary or multi-class classification,
mapping to 
political leanings using e.g.,
Support-Vector Machines, Logistic Regression, and Random Forest techniques \cite{RODRIGOGINES2024121641} with
hand-crafted feature extraction.
\citet{Hube2018} 
adopt a rule-based 
strategy,
defining a list of inflammatory terms
and expanding it with
Word2Vec \cite{Mikolov2013EfficientEO} 
from
Conservapedia articles\footnote{Conservapedia is a wiki-based resource shaped by right-conservative ideas \cite{Hube2018}} to
create a lexicon of politically charged words. 

Our rule-based model 
differs by not relying on lists of
predefined terms;
instead, 
it assumes that differing 
stances towards an entity
across 
articles indicate bias.
This simpler approach
hinges on 
stances towards notable entities, 
differing from the single-sentence approach of \citet{Hube2018}. 
Our model's theoretical foundation suggests
that differences in stance expression
between media outlets signal statement bias.


Bias detection research favors
Transformers over 
Recurrent Neural Networks (RNNs), due
to their self-attention mechanism
for modeling
sequential structures 
\cite{NIPS2017_3f5ee243}. 
However, 
their reliance
on low-level lexical information
\cite{RODRIGOGINES2024121641},
is often insufficient for political bias detection.\footnote{Our
convolutional NN model 
implementation is also affected by this constraint.}
\citet{chen-etal-2020-detecting} attempt to 
overcome 
hand-crafted feature extraction limitations, 
while 
avoiding 
deep learning's pitfalls, by
analyzing second-order information, 
like the frequency and 
order of biased statements, 
and employing
machine-learning methods for bias detection.
Our hybrid approach 
aligns with this, but focuses 
on human interpretable bias features (e.g. stance), 
and favors deep learning over classical machine learning methods
for a more flexible interpretation of these features.

To develop explainability, techniques such as sensitivity analysis and layer-wise relevance propagation (LRP) 
address the black-box nature of deep learning models \cite{samek2017}. This 
explores the limitations of
deep convolutional architectures
by assessing model performance against training and external articles, and identifying differences in the data that correlate with 
performance variations.

\section{Data Querying and Setup}
\label{sec:data}
Both models require first acquiring article data and
correcting imbalances 
to prevent
model bias.

\subsection{Data Sourcing}

The news feed used to implement the models 
in 
our study
is obtained through The Newscatcher API. \footnote{We are granted an educational license intended for research of non-commercial use.}
This 
API provides flexible methods 
for querying data, allowing users to
specify attributes such as news sources, keywords, topics, etc. Both models are premised on the idea that, by exploring outlets with extreme or centrist political biases, three distinct
categories of bias can be 
identified,
establishing 
ground truth.
This allows for assigning 
far-right, center or far-left political leanings
to each 
group of queried articles. 

\begin{figure}
\begin{center}
\includegraphics[width=7.5cm,height=5cm]{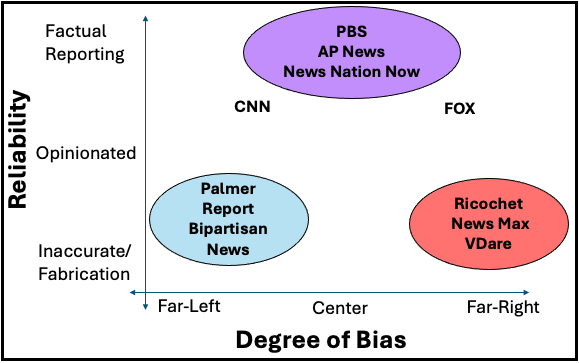}
\end{center}
\vspace*{-.1in}
\caption{News outlet spectrum selected from Media Bias Chart 10.0 \cite{OKC2022}: left, center, right}
\label{Fig:news-outlet-spectrum}
\vspace*{-.25in}
\end{figure}

We first query the available news sources provided by the API and then research political bias charts
to identify trustworthy sources 
and select an eclectic group of news outlets. We adopt a well-known academic media bias classification 10.0 \cite{OKC2022}, which is
based on political bias and reliability.\footnote{See \url{https://adfontesmedia.com/static-mbc/} for a full rendering of news outlets from 2018 through 2024.}
Focusing solely on the political bias dimension, we select 
outlets situated 
within the colored circles
in our simplified rendering of the news outlet spectrum shown in Figure~\ref{Fig:news-outlet-spectrum}. Specifically, \textbf{PBS, AP News, and News Nation Now} are chosen as center outlets, \textbf{Palmer Report and Bipartisan News} 
as far-left outlets, and \textbf{VDare, News Max, and Ricochet} 
as far-right outlets. 

Although it would be 
ideal to consider a greater number of sources for each political category, access to outlets is limited by the 
available
number of API calls 
and 
outlets accessible to the API.
To ensure 
an evenly distributed news feed,
the articles examined are 
restricted
to a three-year range from January 1\textsuperscript{st}, 2021 to December 31\textsuperscript{st}, 2023.

\subsection{Exploratory Data Analysis}
While the query loop aims to collect an equal amount of data from
each outlet, it inadvertently results in
an uneven distribution 
across outlets.\footnote{Appendix ~\ref{Appendix:initial-dist-figures} reveals this discrepancy, showing the distribution by outlet for each three-month period, with certain outlets having significantly more queried articles than others.}
To mitigate
potential model bias,
each category is truncated to only contain ten thousand articles. Additionally, 
due to the depletion of API calls,
this work prioritizes
a balanced distribution 
across all articles relevant
to each class, 
rather than striving for
an equal 
distribution for each outlet. 
By systematically removing
article entries within specific
time intervals for different outlets, the resulting distributions for each category, although
not perfect, 
are substantially improved.\footnote{Appendix ~\ref{Appendix: Training-Data} displays the final state of training data, accomplishing a relatively even distribution across time periods and outlet groupings.}

\section{Model Implementations}
\label{sec:model_implementation}

The rule-based sentiment analysis model isolates sentiment expressed towards both common and proper nouns, leveraging adjectives and verbs
that describe them. This approach aligns with findings from recent research, which 
focus on descriptive language used in relation to specific entities \cite{Alam-Sentiment} to detect bias through sentiment and stance in news articles. The model employs coreference resolution to 
ensure direct reference of verbs and adjectives with correct name entities. Locating the nouns referenced by verbs and adjectives is accomplished through the aforementioned POS reference algorithm, which achieves acceptable performance based on precision, recall and F1 score.


Leveraging these rule-based outputs,
the model creates sentiment vectors embedding the sentiment towards all nouns in articles by political leaning, where each dimension is defined by a unique noun. Sentiment is quantified using the valence scores of all verbs and adjectives considered. It classifies bias by comparing the cosine distance between an article's vector and sentiment vectors for each political leaning. Then the political leaning closest to the article's vector is predicted as its bias.

\begin{figure}
\begin{center}
\includegraphics[width=7.3cm,height=5cm]{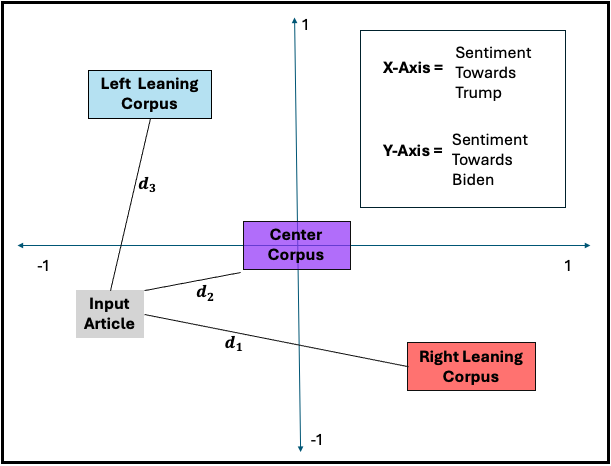}
\end{center}
\caption{ Theoretical mapping of left-vs-right space where an input articles positioned.
}
\label{Fig:Model-visual}
\vspace*{-.2in}
\end{figure}

Figure~\ref{Fig:Model-visual} shows a theoretical mapping of the three corpora projected onto a 2D plane, with
each dimension representing sentiment 
toward a corresponding entity based on
all the adjectives and verbs referencing
it 
within each corpus. As 
expected, the right-leaning corpus shows 
negative sentiment towards Biden and 
positive sentiment towards Trump, while the left-leaning corpus expresses the opposite. The
center-leaning corpus displays a neutral sentiment towards both. An input article is then positioned on the plane based on the sentiment it expresses towards both entities. The shortest cosine distance is found between the input article and the center corpus,
indicating
that the article's stance 
is most 
aligned with the
center corpus. 
This suggests that
a highly negative stance towards Trump, with a moderately negative stance towards Biden, 
indicates 
a politically centered standpoint.

By contrast, the
deep learning model
processes
raw text directly, without 
segmenting or extracting stance-specific meaning.
The convolutional NN model 
captures dependencies and recurring structures in text
through multiple deep learning layers. This model
achieves strong performance (see Section~\ref{sec:result}) 
in classifying articles 
across three outlets. 


Both the rule-based and deep learning models are applied to the preprocessed dataset.
The implementation of the rule-based model (Section~\ref{sec:RB-Sent-Analysis}) is more 
involved than that of the deep-learning convolutional NN model 
(Section~\ref{sec:Convolutional-Model}) in that the latter
uses a standard architecture, whereas the former
proposes a novel design. Our models 
are powered by
2 AMD EPYC 75F3 CPU cores complemented by 2 NVIDIA A100 GPU cores. We use 80\% of the data 
for training and the remaining 20\% 
for testing the models. 
Although the training suite encompasses the full extent of the time period explored, temporal leakage is not an issue, as the models are not devoted to the prediction of bias in future news articles. The idea of a temporal 
dataset is solely meant to provide a more comprehensive span of biased text.

\vspace*{-.03in}
\subsection{Rule-Based Sentiment Analysis Model}
\label{sec:RB-Sent-Analysis}

Our rule-based sentiment analysis implementation 
aims to 
identify political bias by extracting and quantifying the sentiment expressed towards nouns through the verbs and adjectives that refer to
them. This 
involves coreference resolution, dependency parsing, POS reference resolution, sentiment vectorization, and cosine distance as the ultimate classification metric. 
Each step 
is detailed below.


\vspace*{-.03in}
\subsubsection{Coreference Resolution}


Our study resolves coreference to prevent the aggregation of
sentiment for lexically equivalent nouns that 
represent
different entities. 
Two common examples 
in the dataset include the use of pronouns and common nouns to reference
named entities.
For example, the text ``John is gifted. \textit{He} was always good at math.'' becomes
``John is gifted. \textit{John} was always good at math.'' 
This allows us to attribute both the
adjectives ``good'' and ``gifted'' 
to John rather than associating
``gifted'' with ``he''. 

Without coreference resolution,
the resulting sentiment dictionaries would inaccurately average
sentiments expressed toward 
entities 
referred to
by the same pronouns, 
significantly undermining the model's effectiveness.
We employ the spaCy coreference resolution model \cite{coref-res-article},
an end-to-end neural system applicable to various entity coreference tasks.

\subsubsection{Dependency Parsing and Part of Speech Reference Resolution}

With coreferences resolved, the model 
associates verbs and adjectives with their corresponding nouns
using spaCy dependency trees \cite{spacy2}.
The complexity of dependency paths 
makes rule-based 
identification of
all 
noun-adjective-verb relations
challenging.
To maintain 
sentiment accuracy without high computational costs,
we 
balance
relation accuracy with the number of identified relations.

We observe that, while nouns are not always associated with a modifying verb or adjective, verbs and adjectives almost always imply the presence of a noun.
Accordingly, instead
of finding relations 
from nouns to verbs and adjectives, the method 
identifies relations from verbs and adjectives to 
the nearest
noun, regardless of 
its position. This noun is then considered the one being referenced by the verbs and adjectives it stems from. 


The algorithm uses bottom-up dynamic programming to reduce 
complexity. 
It progressively updates (int, string) pairs corresponding to each token in a sentence with the distance from the closest noun to that token and the noun itself with a complexity of $O(N^2)$. 

\begin{figure}
\begin{center}
\vspace*{-.1in}
\includegraphics[width=7cm,height=3.3cm]{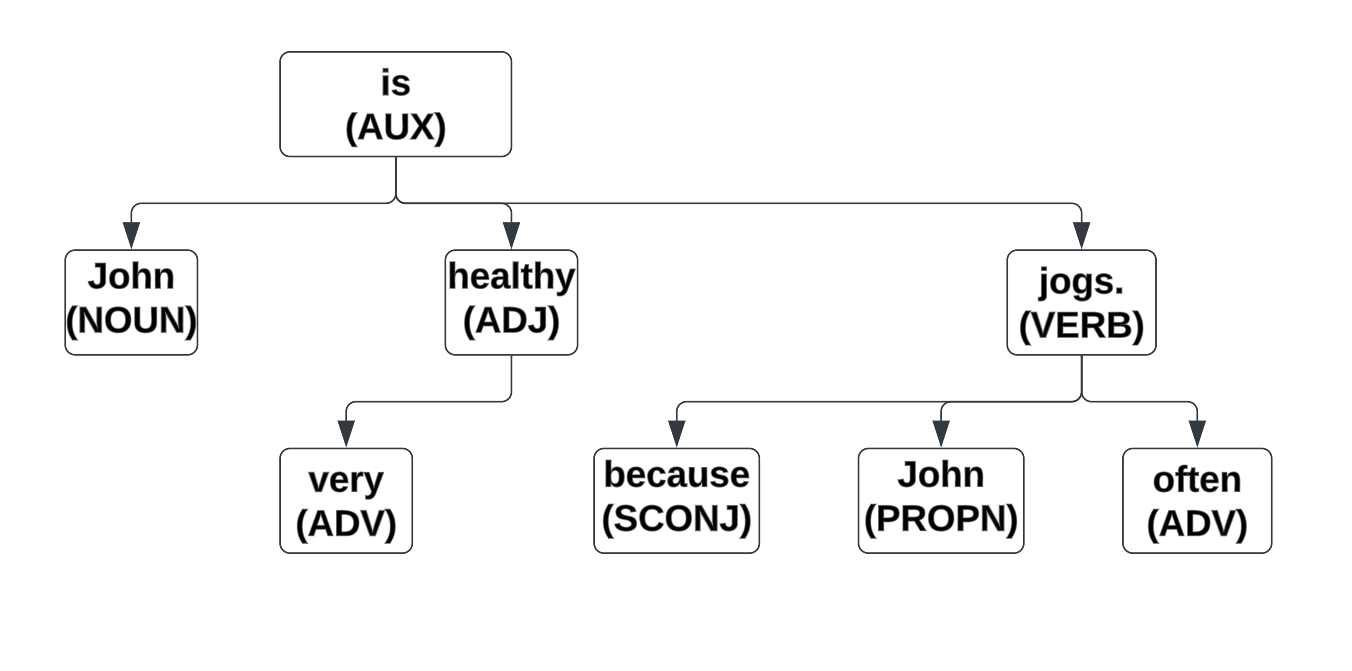}
\end{center}
\vspace*{-.3in}
\caption{Dep Tree: Algorithm Walk-through }
\vspace*{-.2in}
\label{Fig:dep-tree-alg-walkthrough}
\end{figure}

For 
clarity, consider the sentence ``John is very healthy because \textit{he} often jogs'' after
coreference resolution:
``John is very healthy because \textit{John} often jogs''. 
Figure~\ref{Fig:dep-tree-alg-walkthrough} shows the dependency tree for this enriched sentence. A memoization array of length eight is initialized (Table~\ref{tab:mem-alg-walkthrough}). 
For example, 
the 
entry at index 1
contains a distance of 1 
because the noun ``John'' is a child of the auxiliary ``is''. 


\begin{table}[h]
\centering
{\small
\begin{tabularx}{\columnwidth}{|c|X|X|X|X|X|X|X|X|X|}
    \hline
    \textbf{Index} & 0 & 1 & 2 & 3 & 4 & 5 & 6 & 7 \\ \hline
    \textbf{Dist.} & 0 & 1 & -1 & -1 & -1 & 0 & -1 & 1 \\ \hline
    \textbf{Noun}  & John & John &  &  &  & John &  & John \\ \hline
\end{tabularx}
}
\vspace*{-.1in}
\caption{Memoization: Algorithm Walk-through}
\label{tab:mem-alg-walkthrough}
\vspace*{-.15in}
\end{table}

Starting with verbs, the algorithm 
examines index 7 
corresponding to the verb ``jogs'' 
with entry (1, \textquotedbl John\textquotedbl) and
traverses upward $D-1$ times to find a closer distance. Since $D=1$, no parents are considered
and ``jogs'' is associated
with ``John''. A similar procedure is followed for adjectives, e.g., index 3 corresponds to 
``healthy'' 
with entry (-1, \textquotedbl\textquotedbl). Given that there is no noun successor to ``healthy,'' the algorithm 
recursively traverses parent nodes 
to find the shortest distance. 
It considers
the auxiliary ``is'', corresponding
to entry (1, \textquotedbl John\textquotedbl). Since 
``is'' is one edge above ``healthy,'' 
and ``John'' is one edge above ``is,'' ``John'' is determined to be
two edges away from ``healthy.'' 
Further traversal is unnecessary as no shorter distance exists, confirming 
that ``healthy'' corresponds to ``John''.

In order to verify the algorithm’s viability for POS reference resolution, we use a dataset containing 
100 random sentences, 
comparing the algorithm's identified relations with human-annotated relations.
The sentences are generated using  
Chat GPT 4o~\cite{chatgpt_4}, and varied in their use of verbs, adjectives, and nouns. The POS relations used as a ground truth are verified by a human.
The tests yield 
F1-Scores of 0.80 for adjective-noun relations and 0.71 for verb-noun relations.  

\subsubsection{Sentiment Vectorization and Cosine Distance Computation}

Sentiment vectorization 
defines
an $N$-dimensional space 
with a vector of length $N$, where each dimension pertains to a unique noun 
from the corpus of articles. Each group of news outlets 
is mapped onto this 
space. 
An article is mapped 
by extracting the sentiment expressed towards 
nouns in
it, 
considering only those nouns 
present
in the original training corpus. 
The closest vector indicates the primary political leaning expressed in the article.

Using dependency parsing and the POS Reference Resolution algorithm to identify adjective and verb references to nouns, 
we create a mapping that associates
each noun in an article with a list of 
referencing adjectives and verbs.
The valence score of each verb and adjective is extracted using the TextBlob library \cite{loria2018textblob},
where valence
indicates
the degree of positive or negative sentiment.
The sentiment 
towards a noun is defined
by the average valence score of all referring
adjectives and verbs.
This process is repeated
for each article within a political leaning.
By merging these mappings, we obtain final mapping 
with keys representing all unique
nouns in the corpus, each pointing to
the average valence score of all its mentions.

Applying this process to
each of the three article groups 
produces three distinct mappings,
each containing the nouns found in their respective corpus. To compare these groups, the 
mappings are consolidated to hold references to the same group of nouns. 
A separate mapping 
assigns
an index to each noun 
across all three 
corpora. If $N$ distinct nouns are identified,
a vector of length $N$ is defined for each 
article group.
The
sentiment expressed towards $K$th noun is assigned to index $K$. 
For 
indices 
where the relevant noun is absent
in a group’s corpus, a sentiment score of 0 
is assigned.

This process produces three equal-length vectors
in an $N$-dimensional space,
representing 
sentiment toward all nouns 
in the corpus.  
To classify an article, 
valence scores for all nouns are computed
and added onto an $N$-length vector,
ignoring nouns 
not in the 
training corpus. The cosine distance between this vector and each of the three original vectors 
is calculated, assigning
the article the political leaning of the closest vector.

Consider the simplified example in Table~\ref{Fig:noun-sent-corpus}, where three sub-tables show the stance of
each article group towards their respective nouns.
Each unique noun 
is assigned an identifier
(Table~\ref{Fig:noun-ident}). 
%
Using this mapping, 
the 
initial stance
tables are converted into vectors of length 6 (Table~\ref{Fig:vector-process}), with absent
nouns 
assigned a score of 0.




\begin{table}[h]
\centering
\small
\begin{tabularx}{\columnwidth}{|c|l|l|l|X|}
    \hline
    \textbf{Noun} & Trump & IRA & Israel & Immigrant \\ \hline
    \textbf{Left Stan.} & -0.7 & 0.5 & 0.1 & 0.3 \\ \hline
\end{tabularx}
\vspace{1em}

\begin{tabularx}{\columnwidth}{|c|l|l|l|X|}
    \hline
    \textbf{Noun} & Trump & IRA & Israel & Vaccine \\ \hline
    \textbf{Right Stan.} & 0.8 & -0.1 & 0.8 & -0.5 \\ \hline
\end{tabularx}
\vspace{1em}

\begin{tabularx}{\columnwidth}{|c|X|X|X|X|}
    \hline
    \textbf{Noun} & Trump & IRA & Israel & China \\ \hline
    \textbf{Center Stan.} & -0.2 & 0.1 & 0.3 & -0.1 \\ \hline
\end{tabularx}
\vspace*{-.1in}
\caption{Noun Stance by Corpus} 
\label{Fig:noun-sent-corpus}
\vspace*{.1in}
\centering
\small
\begin{tabularx}{\columnwidth}{|c|c|c|c|c|X|}
    \hline
    {Trump} & IRA & Israel & Immigrant & Vaccine & China \\ \hline
    {0} & 1 & 2 & 3 & 4 & 5 \\ \hline
\end{tabularx}
\vspace*{-.1in}
\caption{Noun Identifier Mapping}
\label{Fig:noun-ident}
\vspace*{-.15in}
\end{table}
\begin{table}[h]
\centering
\small
\begin{tabularx}{\columnwidth}{|c|X|X|X|X|X|X|}
    \hline
    \textbf{Identifier} & 0 & 1 & 2 & 3 & 4 & 5 \\ \hline
    \textbf{Left} & -0.7 & 0.5 & 0.1 & 0.3 & 0.0 & 0.0 \\ \hline
    \textbf{Right} & 0.8 & -0.1 & 0.8 & 0.0 & -0.5 & 0.0 \\ \hline
    \textbf{Center} & -0.2 & 0.1 & 0.3 & 0.0 & 0.0 & -0.1 \\ \hline
\end{tabularx}
\vspace*{-.1in}
\caption{Vectorization}
\label{Fig:vector-process}
\vspace*{-.15in}
\end{table}







The first sub-table of Table~\ref{Fig:eval-process} shows a stance dictionary for an article to be classified, listing all nouns 
in the article and 
their associated stance scores.
The second sub-table 
appends this dictionary to the end of
Table~\ref{Fig:vector-process} using the aforementioned index mapping. Note that 
``Canada'', a noun
not in the training corpus, is 
absent from the 
classification vector. 
The third sub-table shows 
classification
by calculating the cosine distance between the article's vector and each of the three vectors 
representing 
political leanings.

\begin{table}[h]
\vspace*{-.05in}
\centering
\small
\begin{tabularx}{\columnwidth}{|c|X|X|X|}
    \hline
    \textbf{Noun} & Trump & Immigrant & Canada \\ \hline
    \textbf{Article Stan.} & -0.3 & 0.10 & 0.05 \\ \hline
\end{tabularx}
\vspace{1em}

\begin{tabularx}{\columnwidth}{|c|X|X|X|X|X|X|}
    \hline
    \textbf{Identifier} & 0 & 1 & 2 & 3 & 4 & 5 \\ \hline
    \textbf{Left} & -0.7 & 0.5 & 0.1 & 0.3 & 0.0 & 0.0 \\ \hline
    \textbf{Right} & 0.8 & -0.1 & 0.8 & 0.0 & -0.5 & 0.0 \\ \hline
    \textbf{Center} & -0.2 & 0.1 & 0.3 & 0.0 & 0.0 & -0.1 \\ \hline
    \textbf{Article} & -0.3 & 0.0 & 0.0 & 0.1 & 0.0 & 0.0 \\ \hline
\end{tabularx}
\vspace{1em}

\begin{tabularx}{\columnwidth}{|c|X|X|X|}
    \hline
    \textbf{X} & Left & Right & Center \\ \hline
    \textbf{Cosine Dist.(Article, X)} & 0.17 & 1.51 & 0.51 \\ \hline
\end{tabularx}
\vspace*{-.1in}
\caption{Evaluation Process}
\label{Fig:eval-process}
\vspace*{-.2in}
\end{table}






\subsection{Convolutional NN Model}
\label{sec:Convolutional-Model}

We choose to use a convolutional NN model to classify bias since convolutional models employ a highly unconstrained assessment of features through their convolutional and pooling layers, which can capture complex patterns in text data. A Convolutional NN is chosen over common models applied to textual analysis (e.g. transformers) for their ability to apply a uniform focus on features across the input, maintaining a more liberal and direct interpretation of data. This unconstrained feature assessment contributes to a lack of explainability, as the internal logic of the convolutional model
remains opaque. In the sections below, we discuss opaqueness as a limitation that challenges 
complete reliance on deep learning methodologies 
for complex classification tasks. Instead, we argue that rule-based or hybrid approaches would provide greater transparency.

Inspired by
the work of \citet{atmosera2023},
we combine 
datasets 
representing three political leanings.
After removing stop words from each article, a Keras tokenizer assigns an index to each unique word enabling the neural network to interpret input and identify patterns for political classification. The input embeddings are of 32 dimensions and the model consists of two convolution layers with a max pooling layer in between and a global max pooling layer at the end. The model is trained over five epochs using the Adam optimizer and categorical cross-entropy loss to improve accuracy using a validation dataset.\footnote{Details regarding training and validation process are provided in Appendix~\ref{Appendix:convolutional-model-training}} We use the tensorflow library under the Apache License 2.0.
(For more details, see 
Appendix~\ref{Appendix:convolutional-model-architecture}.)

\section{Results}
\label{sec:result}
We evaluate both models' performance by examining
precision, recall, and F1 across the three classes. 
,using a dataset comprising 20$\%$ of the original data. Results reflect the models' classification accuracy, not their ability to recognize political bias. We revisit this distinction 
in our evaluation of model performance on external news outlets. 

\subsection{Sentiment Analysis Model: Evaluation}

Each tested article  undergoes
coreference resolution,
and the sentiment 
towards all nouns is quantified into
a vector. This vector is then compared to 
the vectors 
representing
the three political leanings.
The article is assigned the political leaning
corresponding to
the closest 
vector.





Table~\ref{Tab:rule-based-model-metrics} 
shows the performance of the
rule-based model for each classification group.
The \textit{Left} class has the highest F1-Score of 0.57, 
with the \textit{Center} and \textit{Right} classes having slightly lower F1-Scores of 0.51 and 0.52, respectively. 

\begin{table}[H]
\vspace*{-.05in}
\footnotesize{
\begin{tabular}{p{1.5cm}p{1.5cm}p{1.5cm}p{1.5cm}}
\hline
\multicolumn{1}{|p{1.5cm}}{} & 
\multicolumn{1}{|p{1.5cm}}{\centering
\textbf{Precision}} & 
\multicolumn{1}{|p{1.5cm}}{\centering
\textbf{Recall}} & 
\multicolumn{1}{|p{1.5cm}|}{\centering
\textbf{F1-Score}} \\ 
\hline
\multicolumn{1}{|p{1.5cm}}{\centering
\textbf{Left}} & 
\multicolumn{1}{|p{1.5cm}}{\centering
0.78} & 
\multicolumn{1}{|p{1.5cm}}{\centering
0.45} & 
\multicolumn{1}{|p{1.5cm}|}{\centering
0.57} \\ 
\hline
\multicolumn{1}{|p{1.5cm}}{\centering
\textbf{Center}} & 
\multicolumn{1}{|p{1.5cm}}{\centering
0.42} & 
\multicolumn{1}{|p{1.5cm}}{\centering
0.66} & 
\multicolumn{1}{|p{1.5cm}|}{\centering
0.51} \\ 
\hline
\multicolumn{1}{|p{1.5cm}}{\centering
\textbf{Right}} & 
\multicolumn{1}{|p{1.5cm}}{\centering
0.48} & 
\multicolumn{1}{|p{1.5cm}}{\centering
0.56} & 
\multicolumn{1}{|p{1.5cm}|}{\centering
0.52} \\ 
\hline
\end{tabular}
}
\vspace*{-.1in}
\caption{Rule-based Model: Metrics}
\label{Tab:rule-based-model-metrics}
\vspace*{-.15in}
\end{table}

\subsection{Convolutional NN Model: Evaluation}

Table~\ref{Tab:convolutional-metrics} 
shows
the convolutional NN model's performance 
for
each classification group.
The \textit{Left} class has the highest F1-Score of 0.98, indicating excellent performance,
with \textit{Center} and \textit{Right} classes having 
F1-Scores of 0.93 and 0.91, respectively. 

\begin{table}[H]
\vspace*{-.1in}
\footnotesize{
\begin{tabular}{p{1.5cm}p{1.5cm}p{1.5cm}p{1.5cm}}
\hline
\multicolumn{1}{|p{1.5cm}}{} & 
\multicolumn{1}{|p{1.5cm}}{\centering
\textbf{Precision}} & 
\multicolumn{1}{|p{1.5cm}}{\centering
\textbf{Recall}} & 
\multicolumn{1}{|p{1.5cm}|}{\centering
\textbf{F1-Score}} \\ 
\hline
\multicolumn{1}{|p{1.5cm}}{\centering
\textbf{Left}} & 
\multicolumn{1}{|p{1.5cm}}{\centering
0.98} & 
\multicolumn{1}{|p{1.5cm}}{\centering
0.98} & 
\multicolumn{1}{|p{1.5cm}|}{\centering
0.98} \\ 
\hline
\multicolumn{1}{|p{1.5cm}}{\centering
\textbf{Center}} & 
\multicolumn{1}{|p{1.5cm}}{\centering
0.91} & 
\multicolumn{1}{|p{1.5cm}}{\centering
0.96} & 
\multicolumn{1}{|p{1.5cm}|}{\centering
0.93} \\ 
\hline
\multicolumn{1}{|p{1.5cm}}{\centering
\textbf{Right}} & 
\multicolumn{1}{|p{1.5cm}}{\centering
0.94} & 
\multicolumn{1}{|p{1.5cm}}{\centering
0.88} & 
\multicolumn{1}{|p{1.5cm}|}{\centering
0.91} \\ 
\hline
\end{tabular}
}
\vspace*{-.1in}
\caption{Convolutional NN Model: Metrics}
\label{Tab:convolutional-metrics}
\vspace*{-.15in}
\end{table}




\subsection{Application and Insights: CNN and FOX}

The applicability of both models is explored by classifying articles 
from news outlets not included in the training corpus. 
This approach distinguishes
a model’s ability to 
recognize bias 
from simply differentiating between training outlets.
The rule-based approach 
aims to target and extract text features 
that express stance, 
ignoring non-political rhetoric or features.
Conversely, the convolutional NN model 
is allowed complete freedom to differentiate between corpora by any means
available. This makes the convolutional NN model sensitive to corpora that show distinguishing 
features past their expression of political bias. 

Although the convolutional NN model 
accurately categorizes articles in the training corpus, this does not necessarily
translate to 
accurate interpretation of bias. By the same token, the rule-based model's 
lower accuracy 
in classifying
articles does not mean it is worse at recognizing bias than the covolutional NN model.  
To focus solely on 
political bias detection,
we 
exclude test outlets
from
the training corpus, preventing
the models from leveraging
similarities in prose, structure, and other lexical features 
within each group of outlets.

CNN and FOX News are 
used to 
test the models beyond the outlets in the training data.
These outlets are chosen because they are
among
the country’s largest news media corporations and are widely acknowledged 
for representing opposite ends of the political spectrum. 
While their opinions are expected to align closer to
with 
left-leaning and right-leaning classes, they are also widely read and resemble center-leaning articles
in style and structure. We consider 1,500 
articles for each outlet
over the three-year period 
of the training corpus.\footnote{Appendix ~\ref{Appendix:App-Data} illustrates the distribution of CNN and Fox articles, with roughly 1500 articles classified for each outlet}
To incorporate a temporal analysis and evaluate the models' predictions across different periods of political tension, batches for each three-month period within the three years are classified separately. Tables ~\ref{Tab:CNN-App} and ~\ref{Tab:Rule-Based-App} show the distribution of predictions across political leanings throughout each time period for each model applied on both FOX and CNN articles. A darker shade for a given entry indicates a higher percentage of articles classified as pertaining toward that political leaning for that time period. 

\renewcommand{\arraystretch}{1.5}  

\begin{table}[H]
 \Large
  \centering
    \centering
    \resizebox{\linewidth}{!}{
      \begin{tabular}{|p{1.5cm}|p{1.5cm}|p{1.5cm}|p{1.5cm}|p{1.5cm}|p{1.5cm}|p{1.5cm}|p{1.5cm}|p{1.5cm}|p{1.5cm}|p{1.5cm}|}
        \hline
        & 21Q2 & 21Q3 & 21Q4 & 22Q1 & 22Q2 & 22Q3 & 22Q4 & 23Q1 & 23Q2 & 23Q3 \\
        \hline
        Left & \cellcolor{blue!15}0.09 & \cellcolor{blue!18}0.09 & \cellcolor{blue!4}0.02 & \cellcolor{blue!2}0.01 & \cellcolor{blue!8}0.04 & \cellcolor{blue!12}0.06 & \cellcolor{blue!14}0.07 & \cellcolor{blue!2}0.01 & \cellcolor{blue!14}0.07 & \cellcolor{blue!14}0.07 \\
        \hline
        Center & \cellcolor{violet!62}0.62 & \cellcolor{violet!77}0.67 & \cellcolor{violet!89}0.79 & \cellcolor{violet!91}0.81 & \cellcolor{violet!93}0.83 & \cellcolor{violet!80}0.7 & \cellcolor{violet!85}0.75 & \cellcolor{violet!83}0.73 & \cellcolor{violet!82}0.62 & \cellcolor{violet!71}0.61 \\
        \hline
        Right & \cellcolor{red!58}0.29 & \cellcolor{red!48}0.24 & \cellcolor{red!38}0.19 & \cellcolor{red!36}0.18 & \cellcolor{red!26}0.13 & \cellcolor{red!48}0.24 & \cellcolor{red!36}0.18 & \cellcolor{red!50}0.25 & \cellcolor{red!62}0.31 & \cellcolor{red!62}0.31 \\
        \hline
      \end{tabular}
    }
  \hfill
    \centering
            \vspace*{0.05in}
    \resizebox{\linewidth}{!}{
      \begin{tabular}{|p{1.5cm}|p{1.5cm}|p{1.5cm}|p{1.5cm}|p{1.5cm}|p{1.5cm}|p{1.5cm}|p{1.5cm}|p{1.5cm}|p{1.5cm}|p{1.5cm}|}
        \hline
        & 21Q2 & 21Q3 & 21Q4 & 22Q1 & 22Q2 & 22Q3 & 22Q4 & 23Q1 & 23Q2 & 23Q3 \\
        \hline
        Left & \cellcolor{blue!30}0.15 & \cellcolor{blue!42}0.21 & \cellcolor{blue!36}0.18 & \cellcolor{blue!40}0.2 & \cellcolor{blue!54}0.27 & \cellcolor{blue!42}0.21 & \cellcolor{blue!42}0.21 & \cellcolor{blue!48}0.24 & \cellcolor{blue!44}0.22 & \cellcolor{blue!52}0.26 \\
        \hline
        Center & \cellcolor{violet!67}0.57 & \cellcolor{violet!80}0.6 & \cellcolor{violet!75}0.55 & \cellcolor{violet!36}0.26 & \cellcolor{violet!48}0.38 & \cellcolor{violet!60}0.5 & \cellcolor{violet!46}0.36 & \cellcolor{violet!51}0.41 & \cellcolor{violet!57}0.47 & \cellcolor{violet!56}0.46 \\
        \hline
        Right & \cellcolor{red!56}0.28 & \cellcolor{red!38}0.19 & \cellcolor{red!54}0.27 & \cellcolor{red!100}0.54 & \cellcolor{red!72}0.36 & \cellcolor{red!58}0.29 & \cellcolor{red!86}0.43 & \cellcolor{red!70}0.35 & \cellcolor{red!62}0.31 & \cellcolor{red!54}0.27 \\
        \hline
      \end{tabular}
    }
  \caption{Convolutional Model. Bias Classification of
FOX (top) and CNN (bottom) articles Over Time.}
  \label{Tab:CNN-App}
\end{table}
\vspace*{-0.2in}
\begin{table}[H]
 \Large
  \centering
    \centering
    \resizebox{\linewidth}{!}{
      \begin{tabular}{|p{1.5cm}|p{1.5cm}|p{1.5cm}|p{1.5cm}|p{1.5cm}|p{1.5cm}|p{1.5cm}|p{1.5cm}|p{1.5cm}|p{1.5cm}|p{1.5cm}|}
        \hline
        & 21Q2 & 21Q3 & 21Q4 & 22Q1 & 22Q2 & 22Q3 & 22Q4 & 23Q1 & 23Q2 & 23Q3 \\
        \hline
        Left & \cellcolor{blue!20}0.1 & \cellcolor{blue!22}0.11 & \cellcolor{blue!8}0.04 & \cellcolor{blue!22}0.11 & \cellcolor{blue!16}0.08 & \cellcolor{blue!26}0.13 & \cellcolor{blue!22}0.11 & \cellcolor{blue!30}0.15 & \cellcolor{blue!14}0.07 & \cellcolor{blue!18}0.09 \\
        \hline
        Center & \cellcolor{violet!59}0.49 & \cellcolor{violet!65}0.55 & \cellcolor{violet!72}0.62 & \cellcolor{violet!59}0.49 & \cellcolor{violet!65}0.55 & \cellcolor{violet!71}0.61 & \cellcolor{violet!67}0.57 & \cellcolor{violet!70}0.6 & \cellcolor{violet!65}0.55 & \cellcolor{violet!77}0.67 \\
        \hline
        Right & \cellcolor{red!80}0.4 & \cellcolor{red!66}0.33 & \cellcolor{red!66}0.33 & \cellcolor{red!78}0.39 & \cellcolor{red!74}0.37 & \cellcolor{red!52}0.26 & \cellcolor{red!64}0.32 & \cellcolor{red!50}0.25 & \cellcolor{red!76}0.38 & \cellcolor{red!50}0.25 \\
        \hline
      \end{tabular}
    }
  \hfill
    \centering
        \vspace*{0.05in}
    \resizebox{\linewidth}{!}{
      \begin{tabular}{|p{1.5cm}|p{1.5cm}|p{1.5cm}|p{1.5cm}|p{1.5cm}|p{1.5cm}|p{1.5cm}|p{1.5cm}|p{1.5cm}|p{1.5cm}|p{1.5cm}|}
        \hline
        & 21Q2 & 21Q3 & 21Q4 & 22Q1 & 22Q2 & 22Q3 & 22Q4 & 23Q1 & 23Q2 & 23Q3 \\
        \hline
        Left & \cellcolor{blue!8}0.04 & \cellcolor{blue!10}0.05 & \cellcolor{blue!18}0.09 & \cellcolor{blue!32}0.16 & \cellcolor{blue!18}0.09 & \cellcolor{blue!20}0.1 & \cellcolor{blue!12}0.06 & \cellcolor{blue!18}0.09 & \cellcolor{blue!18}0.09 & \cellcolor{blue!16}0.08 \\
        \hline
        Center & \cellcolor{violet!83}0.73 & \cellcolor{violet!80}0.7 & \cellcolor{violet!71}0.61 & \cellcolor{violet!63}0.53 & \cellcolor{violet!76}0.66 & \cellcolor{violet!76}0.66 & \cellcolor{violet!82}0.72 & \cellcolor{violet!79}0.69 & \cellcolor{violet!75}0.65 & \cellcolor{violet!81}0.71 \\
        \hline
        Right & \cellcolor{red!48}0.24 & \cellcolor{red!48}0.24 & \cellcolor{red!6}0.03 & \cellcolor{red!64}0.32 & \cellcolor{red!50}0.25 & \cellcolor{red!50}0.25 & \cellcolor{red!46}0.23 & \cellcolor{red!44}0.22 & \cellcolor{red!52}0.26 & \cellcolor{red!42}0.21 \\
        \hline
      \end{tabular}
    }
  \caption{Rule-Based Model. Bias Classification of FOX
(top) and CNN (bottom) articles Over Time.}
  \label{Tab:Rule-Based-App}
\end{table}

The Convolutional and Rule-Based Model results on FOX articles show that the models often classify most articles in each period as Center-leaning, with the majority of the remaining portion classified as Right-leaning. The Convolutional Model is less likely to predict the Left class compared to the Rule-Based Model, favoring more confident Center predictions. Meanwhile, the Rule-Based Model assigns more articles to the Right class than the Convolutional Model. Both models' predictions lean more toward the Center than public perception of FOX, though they still align with its center-right reputation.

In contrast, the Convolutional Model's results on CNN articles differ from the general perception of the outlet. Over a third of the articles are classified as Center-leaning, with the rest slightly favoring the Right class. The Rule-Based Model classifies most CNN articles as Center, with the remaining majority leaning Right. Although neither model's predictions match CNN's center-left stance, the Rule-Based Model tends to classify articles further left than the Convolutional Model.

\subsection{Model Explainability}

%

The Rule-Based Model's transparency and strong theoretical foundation allow us to attribute shortcomings in both the corpus and external dataset results to specific components of the model architecture. 
A combination of factors prevents the model from accurately assigning sentiment to entities, resulting in instances where detected sentiments do not align with political biases.

\begin{figure}[h]
\begin{center}
\includegraphics[width=0.47\textwidth]{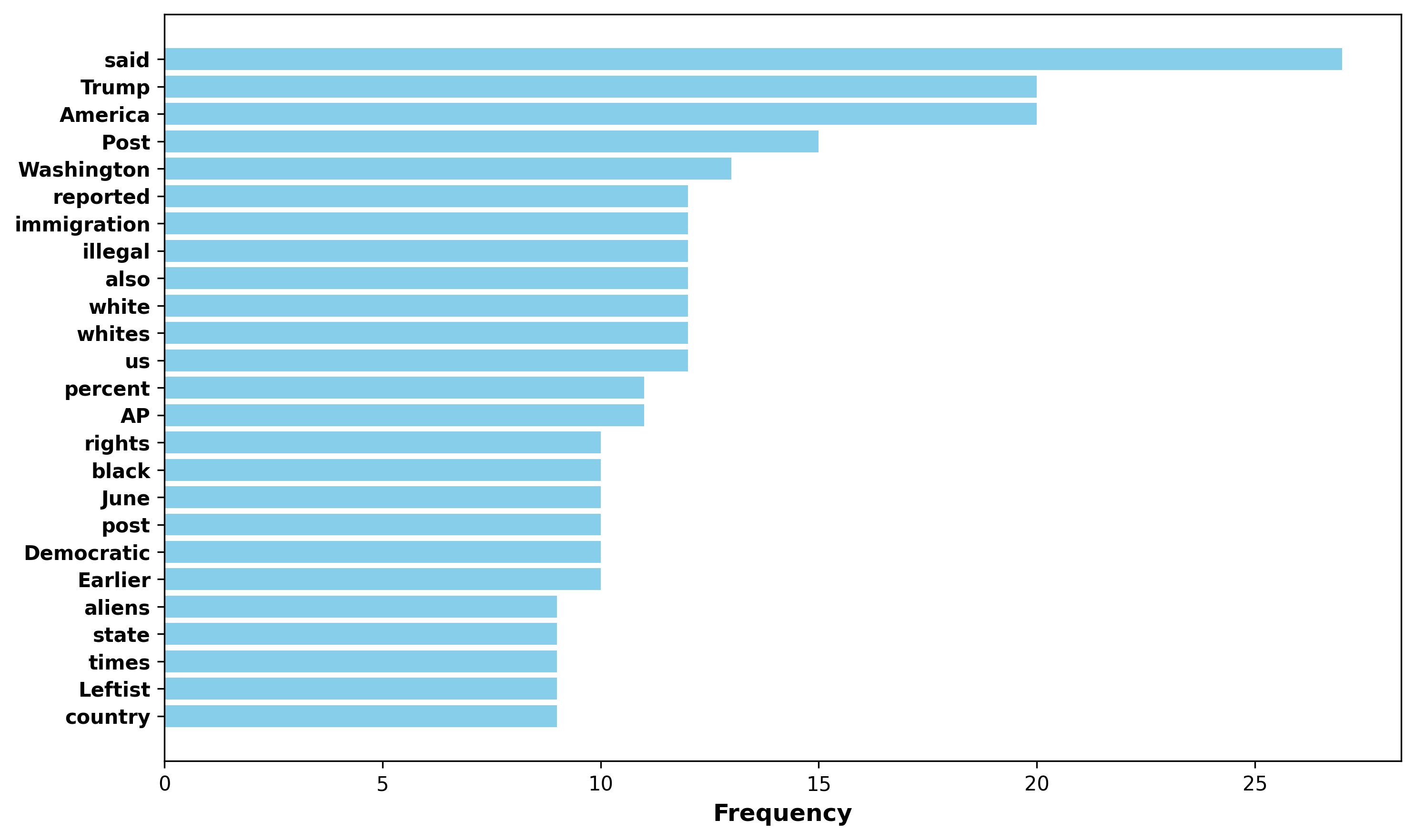}
\end{center}
\vspace*{-.2in}
\caption{Top 25 Influential Terms in Right Class Classification}
\label{Fig:right_leaning_words}
\vspace*{-.2in} 
\end{figure}

\begin{figure}[h]
\begin{center}
\vspace*{-.1in}
\includegraphics[width=0.47\textwidth]{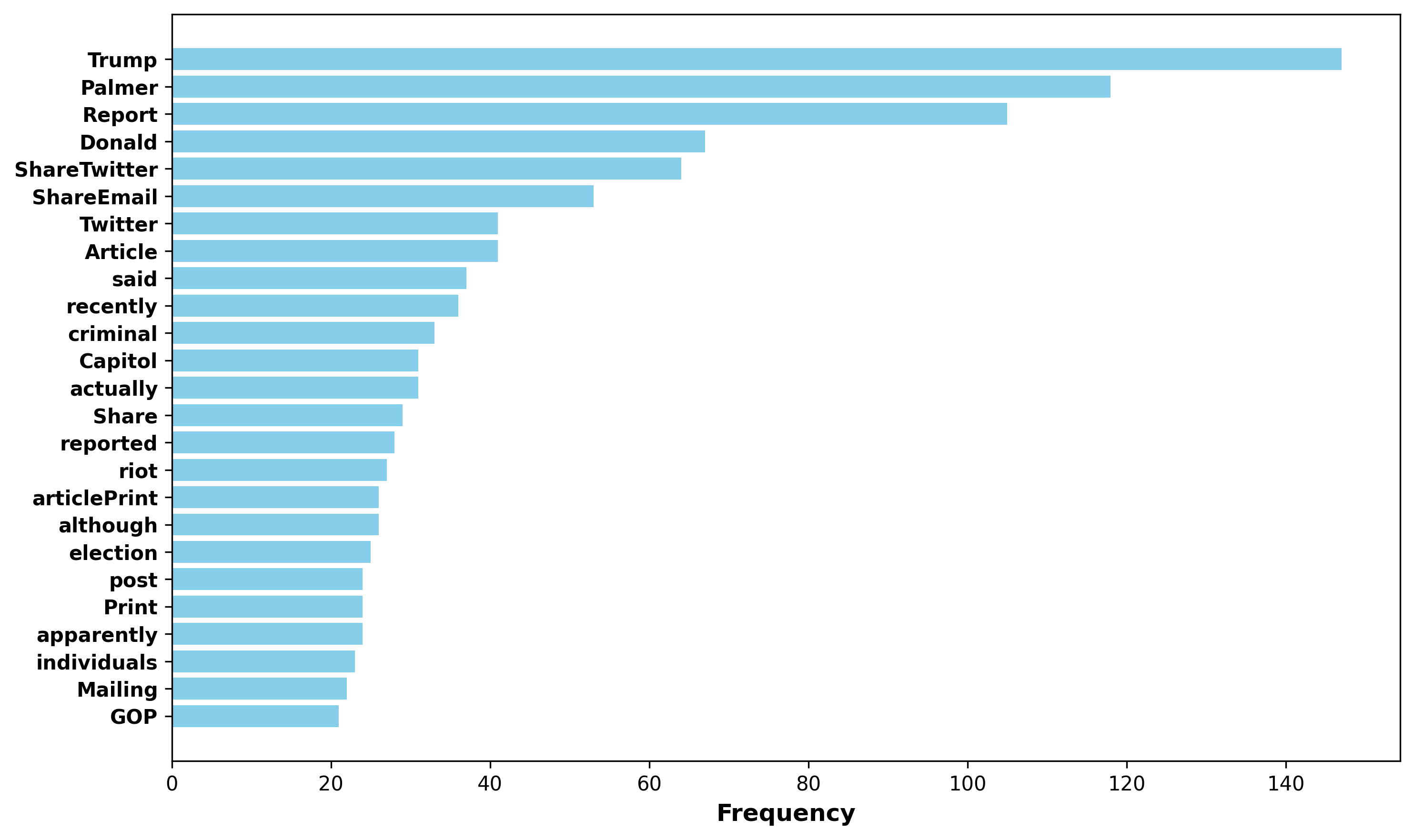}
\end{center}
\vspace*{-.2in}
\caption{Top 25 Influential Terms in Left Class Classification}
\label{Fig:left_leaning_words}
\vspace*{-.1in} 
\end{figure}

\begin{figure}[h]
\begin{center}
\vspace*{-.1in}
\includegraphics[width=0.47\textwidth]{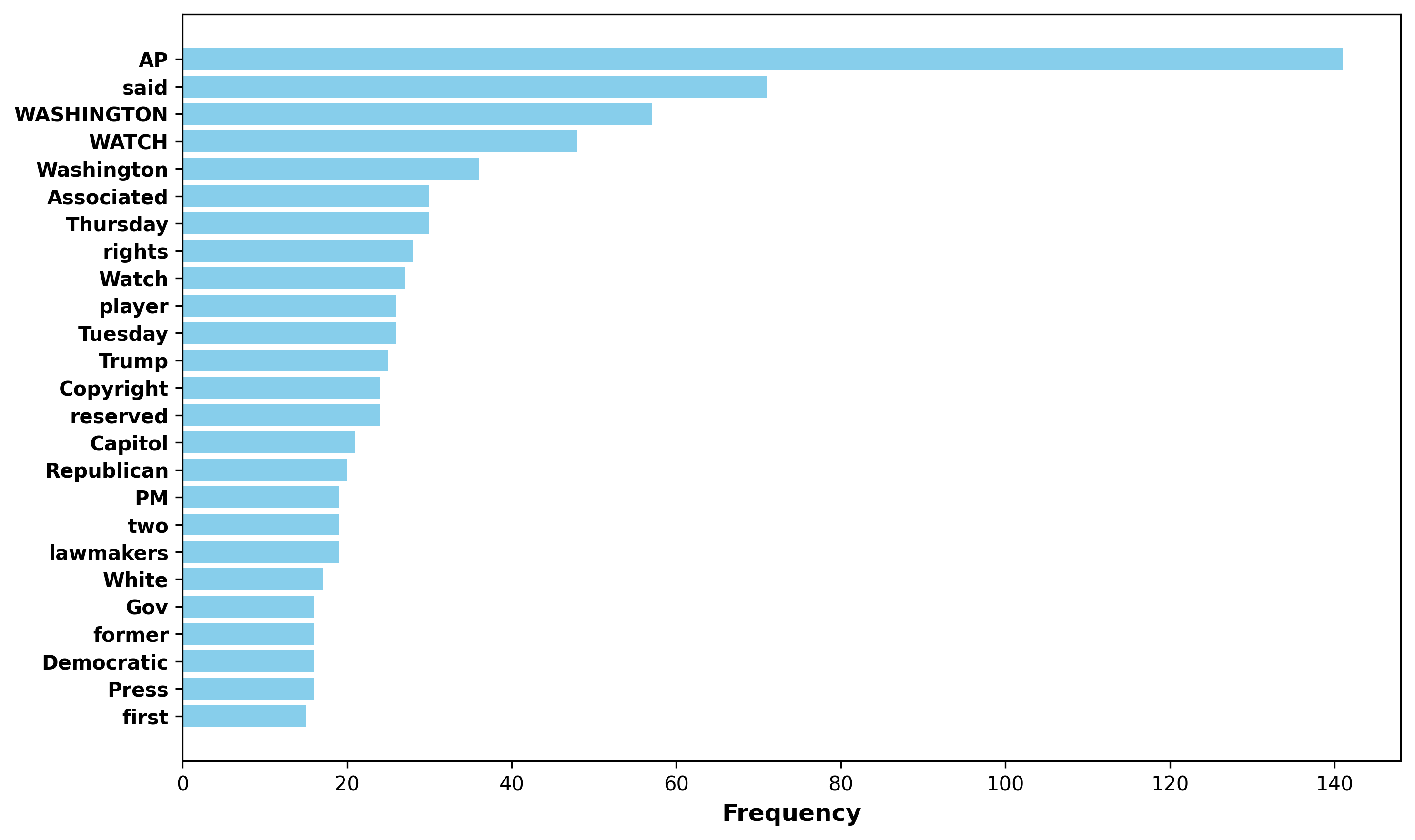}
\end{center}
\vspace*{-.2in}
\caption{Top 25 Influential Terms in Center Class Classification}
\label{Fig:center_leaning_words}
\vspace*{-.2in} 
\end{figure}

Understanding the performance differences between the corpus and external dataset for the Convolutional Model is more challenging due to its black-box deep learning architecture. To investigate this gap, we employ LIME \cite{ribeiro2016whyitrustyou} to identify the words that most influence article classification. A subset of the test suite is analyzed, revealing the 20 most important words in each article's classification. The frequency of the top 25 influential words for each political leaning is shown in Figures ~\ref{Fig:right_leaning_words}, ~\ref{Fig:left_leaning_words}, and ~\ref{Fig:center_leaning_words}. The LIME analysis reveals three types of words that the Convolutional Model relies on for classification.

The first 
influential word types are those that recur frequently due to the limited number of outlets comprising the corpus, e.g.,
``Palmer'' and ``Report'' (Figure~\ref{Fig:left_leaning_words}) and ``AP'' and ``Associated'' (Figure~\ref{Fig:center_leaning_words}).
The second type comprises words that lack political meaning on their own but are common in the rhetoric of certain outlets, e.g., 
``us'' and ``said'' 
(Figure~\ref{Fig:right_leaning_words}) and ``apparently'' 
(Figure~\ref{Fig:left_leaning_words}). 
The third type includes nouns with inherent political connotations, e.g.,
``Trump'', ``Leftist'', ``aliens'' 
``riot'' and ``GOP'' 
(Figure~\ref{Fig:left_leaning_words}) and ``Republican'', ``Democratic'' and ``Capitol'' 
(Figure ~\ref{Fig:center_leaning_words}). 

From this analysis, we conclude that the Convolutional Model struggles to maintain its high performance when applied to external news outlets, primarily because CNN and FOX articles lack the first and second types of influential words that are specific to the outlets used for training. Without relying on rhetoric similarities or outlet-specific names, the model assesses politically charged terms—words it does not emphasize adequately during training to draw reliable conclusions. 

Notably, Right leaning predictions are the least reliant on the first and second type of words, and are generally less reliant on any given word in the classification of articles. This explains why FOX article predictions by the Convolutional model aligns more with the outlet's political stance than CNN article predictions. The improved performance resulting from the Convolutional Models's focus on 
politically charged words supports 
the Rule-Based model's framework, which is primarily designed to detect sentiment towards such words.

Left leaning predictions rely heavily on the first type of influential terms, causing the Convolutional Model to perform poorly on external data when predicting CNN articles. We hypothesize that the superior performance of the Rule-Based Model in this task stems from its ability to focus on nouns associated with political entities, which the Convolutional Model does not sufficiently emphasize in its classification of Left leaning articles. 

\section{Conclusion and Future Work}
\label{sec:conclusion}
This paper examines two models for classifying political bias in news media: a sentiment analysis rule-based model and a convolutional NN model. Given the complexity of politically biased text, a ground truth is established using the political spectrum placement of widely read news outlets by credible academic sources \cite{OKC2022}. 

The rule-based model applies coreference resolution and a POS reference algorithm to extract stance towards nouns, mapping them into an $N$-dimensional space for comparison with input articles. The convolutional NN model focuses on identifying distinctive patterns. 

Results initially indicate that the convolutional NN model significantly outperforms the rule-based in accuracy. However, when models are tested on external data, using CNN and FOX articles, the limitations of the convolutional NN model are uncovered through its significant change in performance. The rule-based model, in contrast, retains its initial performance,
thereby demonstrating its adaptability 
to different datasets.

Potential improvements to the rule-based model include incorporating machine learning techniques for feature extraction and input classification, such as using a decision tree instead of using a closest POS resolution algorithm to indentify noun relations. Alternatively, stance detection would benefit from leveraging more accurate pre-trained models in Aspect Based Sentiment Analysis \cite{hoang-etal-2019-aspect} to better quantify targeted sentiment. Additionally, understanding synonymy through tools like Word2Vec could help map nouns in input articles to similar counterparts in classification vectors, enabling more effective classification. 

Improvements to the Convolutional NN model include prioritization of
explainability and generalizability. Future work involves a thorough data selection process paired with an iterative analysis, using LIME or SHAP, to ensure the use of corpora that do not allow models to hinge predictions on terms unrelated to the classification task. In the classification of bias, this process translates to  
the
prioritization of true bias indicators and disregard of
irrelevant stylistic nuances.

Lastly, future work could 
incorporate
large language models into the study by exploring their performance in the classification of bias and their potential improvement through hybridization techniques. In addition, expanding
the study to include
a training corpus from a diverse range of 
news outlets would help to prevent models from relying on stylistic differences in writing.

\section*{Limitations: A Case for Hybridization}

Overall, this exploration examines the 
extremes of techniques used for media bias classification. It contrasts a clearly defined, rule-based model with a
deep learning model 
that has an opaque internal methodology. The rule-based model, while theoretically sound, 
fails to beat the convolutional NN
in testing, but shows a similar performance in external applications. Both approaches 
have shortcomings that could 
be mitigated through hybridization.  

%
Both models  
are evaluated 
using
ground truth for
political bias 
in news articles,
determined by the publishing outlets 
and 
academic sources that classify
the outlet's political leaning.
However, political bias is 
a highly dynamic,  nuanced, and subjective expression 
that cannot be fully captured
through the perspectives of 
various 
news outlets. While 
our research
aims to investigate 
bias in text, the models we 
construct
are ultimately
designed to classify articles 
based on lexical and syntactic features of the
three corpora considered. 
Thus, they 
classify text, rather than  
classify bias directly. 
Although the dataset 
facilitates temporal and 
diverse analysis of political news media, datasets with articles
annotated specifically for bias would 
provide a more robust ground truth. 
Additionally, the dataset encompasses only US news outlets, which limits the broader international applicability of models trained using it. 

The rule-based sentiment detection model 
focuses solely 
on the sentiment expressed toward nouns, avoiding
irrelevant textual features 
related to political leaning. 
While this approach offers insight into 
how political bias is conveyed,
the model does not target other forms of bias (e.g. Framing bias) 
and additionally considers nouns of a non-political nature 
in its classification process, which may not necessarily indicate
political leaning. 
Beyond its broad 
interpretation of bias, the model faces challenges regarding its practicality.
Since
the model only interprets one feature,
sentiment expressed towards nouns, an article must contain mentions of nouns found
in the corpus for its political leaning 
to be accurately classified. 
Furthermore, if the nouns within the article 
are apolitical or 
rare,
the identified bias 
may lack substantial basis.

The POS reference method
for the rule-based model  
sometimes misses
correct relationships 
or incorrectly identifies them. 
This is because
the algorithm assumes
a one-to-many relationship between nouns and their referencing parts of speech (verbs or adjectives), 
even though many-to-many or many-to-one relationships are possible.  
For example, in the sentence ``John is happy and excited'', 
the
one-to-many relationship between 
the noun ``John'' and the
adjectives ``happy'' and ``excited'' is identified correctly. 
However, in the sentence ``John and Peter are happy'', which has a many-to-one relationship between the
nouns ``John'' and ``Peter'' and the adjective ``happy,'' the algorithm only 
links
``happy'' to
the closest noun, ``Peter''. 

Despite 
the convolutional NN model's
impressive
classification 
performance when tested on outlets found in the training corpus, its focus on political bias as a deciding factor is shown to be insufficient.  
The model accurately categorizes the three classes in
the training corpus, but it 
identifies a strong moderate leaning for FOX and fails to converge on a general political leaning for CNN articles. 
Due to the inherent opacity of 
deep learning models, 
the specific textual features used for classification 
are unpredictable, leaving
developers
to speculate on the mix of features driving 
article classification 
and 
how much
these features are 
influenced by the political bias 
of each outlet. 

Ideally, classification of political bias in news media would 
combine the feature extraction 
of
a rule-based model with the 
self-correction 
of a convolutional NN model. By examining additional text features
that signal
political bias and quantifying
them 
similarly 
to
sentiment expression, a
suitable input vector 
for convolution could be 
generated.
Although the internal processes of the convolutional NN
would remain opaque to the developer,
its 
predictions would focus solely 
on factors 
related to political bias.
Allowing
the developer to set the initial parameters of the neural network
would enable the imposition of constraints
while preserving its self-learning ability,
thereby ensuring that only relevant
resources 
are used for learning.

\section*{Ethics Statement}

The data used for this study is obtained using the News Catcher API, but is otherwise publicly accessible. The API is employed to allow for fast and efficient sourcing of a large number of articles. The integrity of the data is maintained by verifying the reputability of the API used and by assessing the articles queried. 


Maintaining
objectivity is crucial in this study 
on automatic detection of political bias in text. Both implemented models use standardized datasets and transparent processes to ensure a fair analysis of results. 
It is important to emphasize
that our models' evaluation of CNN/FOX bias 
is not intended
as a definitive judgment of their political leaning. Rather, it serves as an 
exercise to demonstrate the capabilities and limitations of NLP techniques in analyzing political bias with respect to a well-known academic media bias classification \cite{OKC2022}.

The use of AI in this study, seen primarily through the CNN model, is done responsibly. We acknowledge AI's limitations in assessing a highly subjective and sensitive subject as is political bias. In fact, this study argues for greater transparency to transcend
opaque deep learning systems.

\section*{Acknowledgement}

This work is supported, in part, by DARPA Contract No. HR001121C0186. Any opinions, findings and conclusions or recommendations expressed in this material are those of the authors and do not necessarily reflect the views of the US Government.



\bibliography{anthology,custom}
\bibliographystyle{acl_natbib}

\appendix

\newpage

\section{Appendix}

\subsection{Initial Data Distribution Figures}
\label{Appendix:initial-dist-figures} 

\begin{figure}[H]
\begin{center}
\includegraphics[width=8cm,height=5cm]{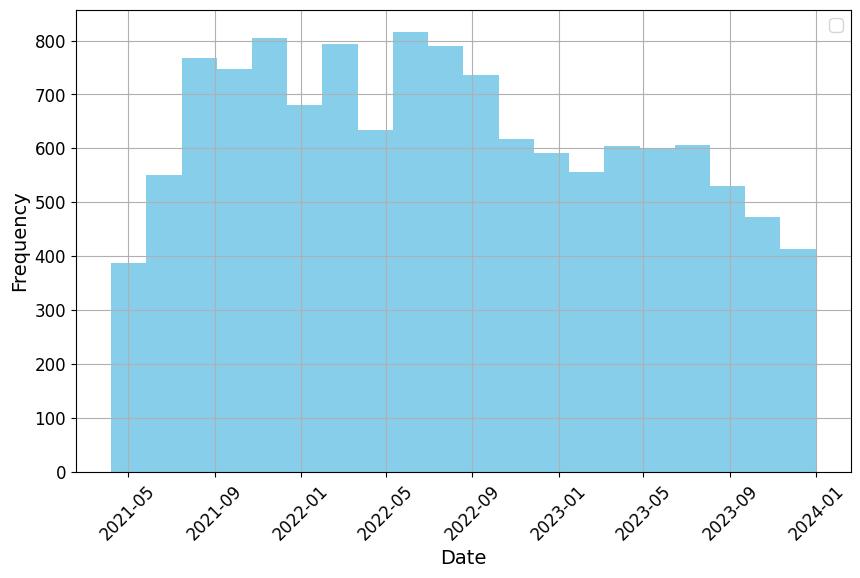}
\end{center}
\vspace*{-.2in}
\caption{Left-Leaning Outlets: Bipartisan News}
\label{Fig:bipartisan}
\end{figure}

\begin{figure}[H]
\begin{center}
\includegraphics[width=8cm,height=5cm]{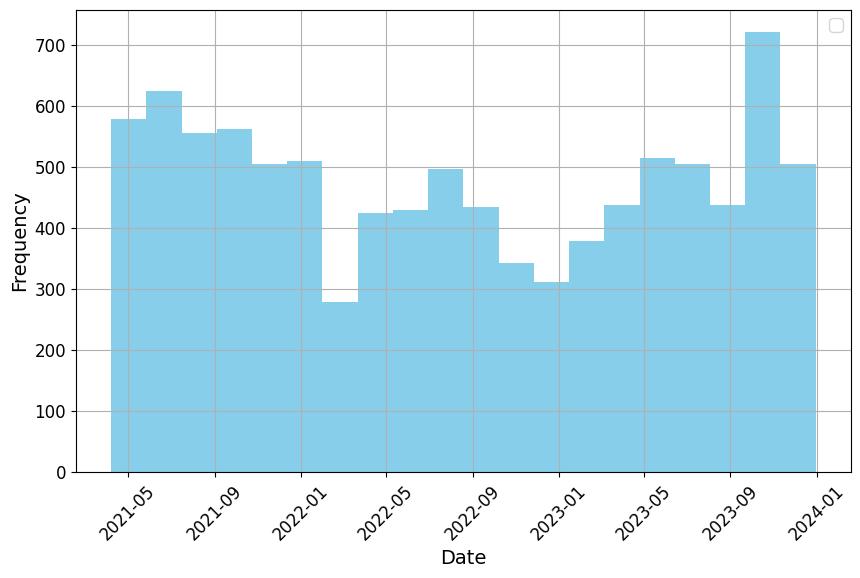}
\end{center}
\vspace*{-.2in}
\caption{Left-Leaning Outlets: Palmer Report}
\label{Fig:Palmer}
\end{figure}

\begin{figure} [H]
\begin{center}
\includegraphics[width=8cm,height=5cm]{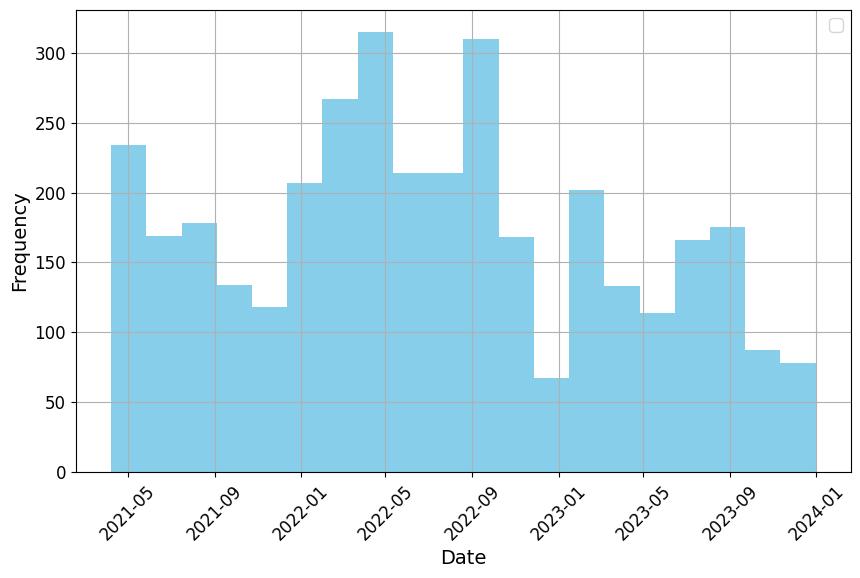}
\end{center}
\vspace*{-.2in}
\caption{Right-Leaning Outlets: VDare }
\label{Fig:VDare}
\end{figure}

\begin{figure}[H]
\begin{center}
\includegraphics[width=8cm,height=5cm]{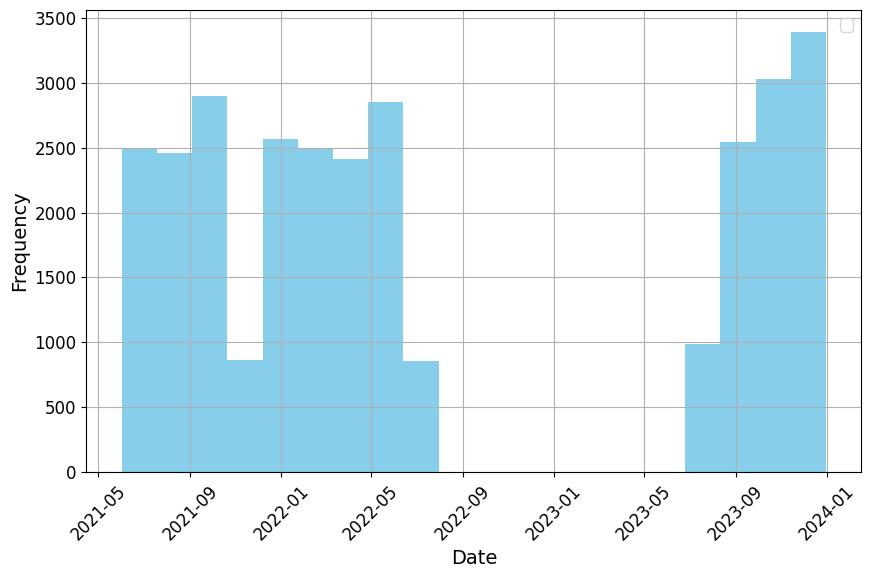}
\end{center}
\vspace*{-.2in}
\caption{Right-Leaning Outlets: News Max}
\label{Fig:NewsMax}
\end{figure}

\begin{figure}[H]
\begin{center}
\includegraphics[width=8cm,height=5cm]{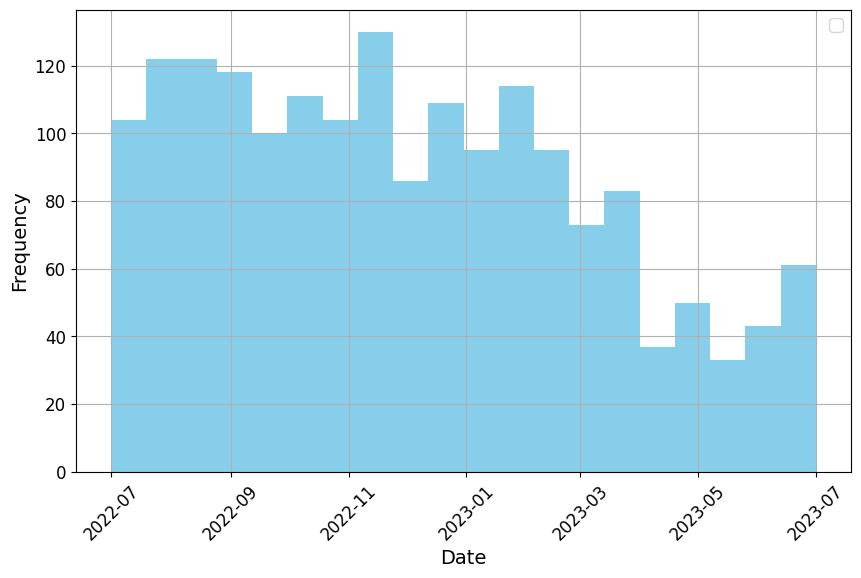}
\end{center}
\vspace*{-.2in}
\caption{Right-Leaning Outlets: Ricochet}
\label{Fig:Ricochet}
\end{figure}

\begin{figure}[H]
\begin{center}
\includegraphics[width=8cm,height=5cm]{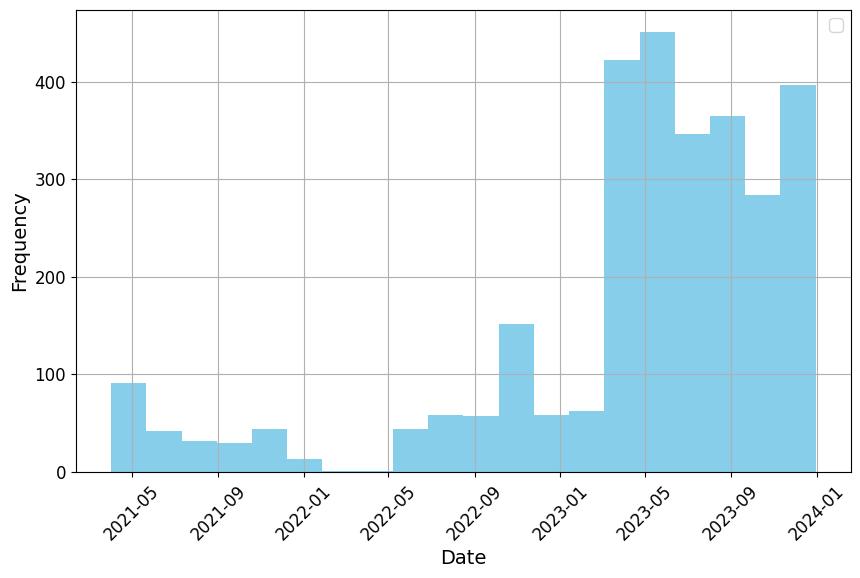}
\end{center}
\vspace*{-.2in}
\caption{Center-Leaning Outlets: News Nation Now}
\label{Fig:NewsNationNow}
\end{figure}

\begin{figure}[H]
\begin{center}
\includegraphics[width=8cm,height=5cm]{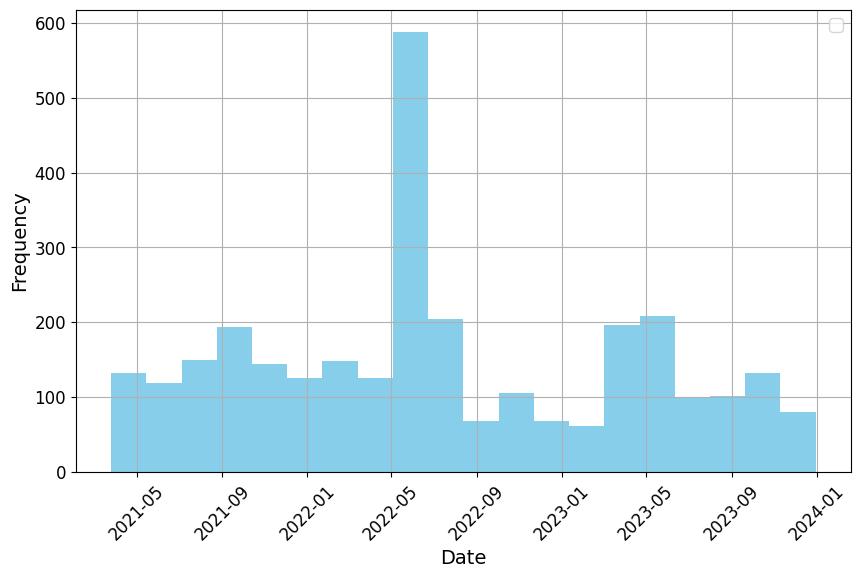}
\end{center}
\vspace*{-.2in}
\caption{Center-Leaning Outlets: AP News}
\label{Fig:APNews}
\end{figure}

\begin{figure}[H]
\begin{center}
\includegraphics[width=8cm,height=5cm]{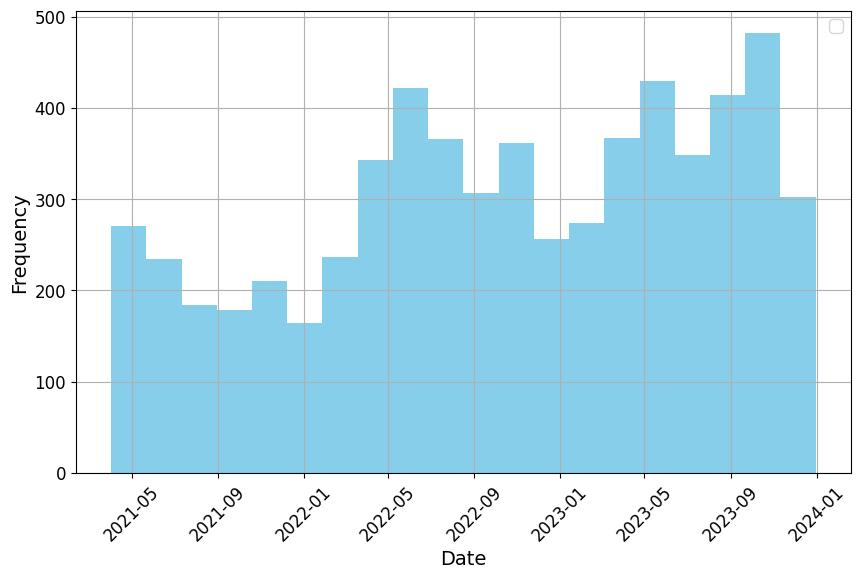}
\end{center}
\vspace*{-.2in}
\caption{Center-Leaning Outlets: PBS}
\label{Fig:PBS}
\end{figure}

Figures~\ref{Fig:bipartisan}--\ref{Fig:PBS} show
the initial distributions for each of the eight news outlets considered in the construction of
both models. The distributions are separated into groups 
based on the 
three biases being explored.

\vspace*{-.3in}

\subsection{Resulting Distribution Across News Outlets for Each Political Grouping}
\label{Appendix: Training-Data}
\begin{figure}[H]
\begin{center}
\includegraphics[width=7.69cm,height=5.03cm]{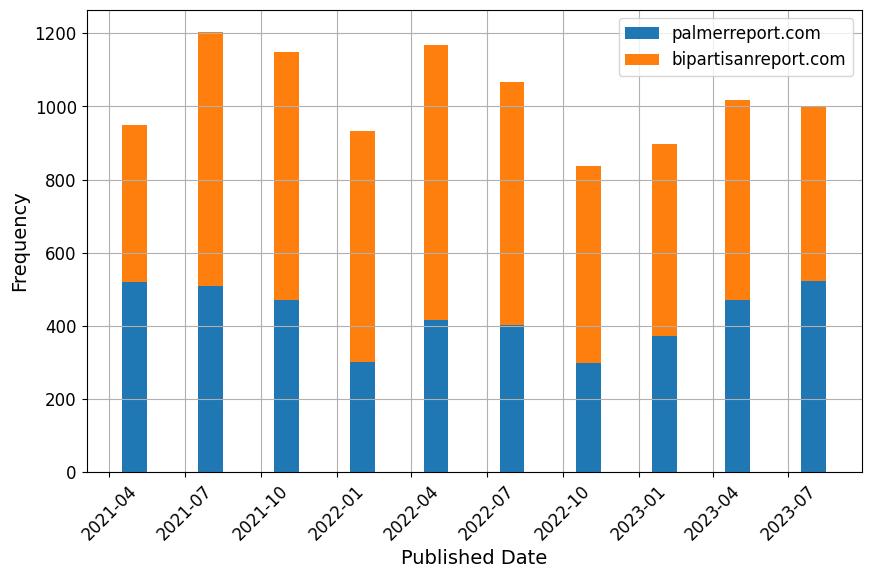}
\end{center}
\vspace*{-.2in}
\caption{Left Outlets Aggregate Distribution}
\label{Fig:left-outlets-distro}
\end{figure}

\begin{figure}[H]
\begin{center}
\includegraphics[width=7.69cm,height=5.03cm]{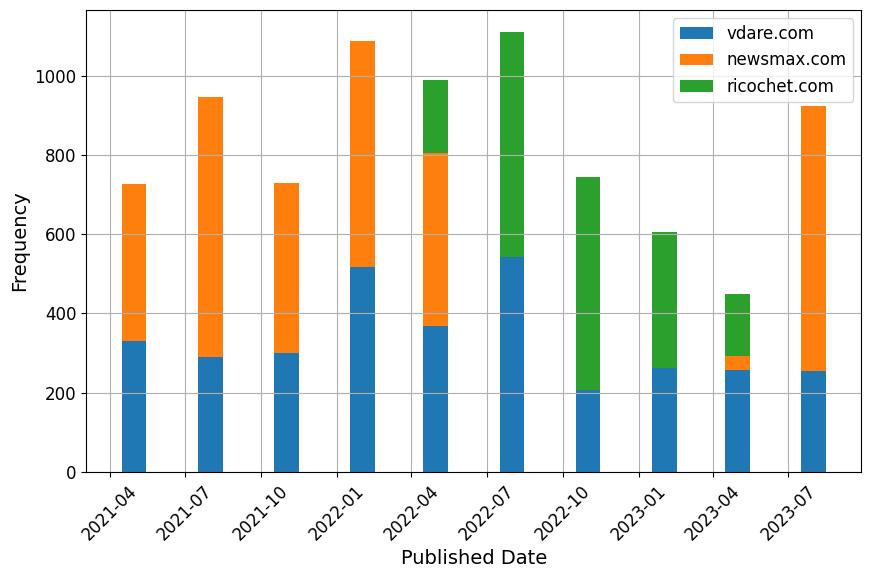}
\end{center}
\vspace*{-.2in}
\caption{Right Outlets Aggregate Distribution}
\label{Fig:right-outlets-distro}
\end{figure}

\begin{figure}[H]
\begin{center}
\includegraphics[width=7.69cm,height=5.03cm]{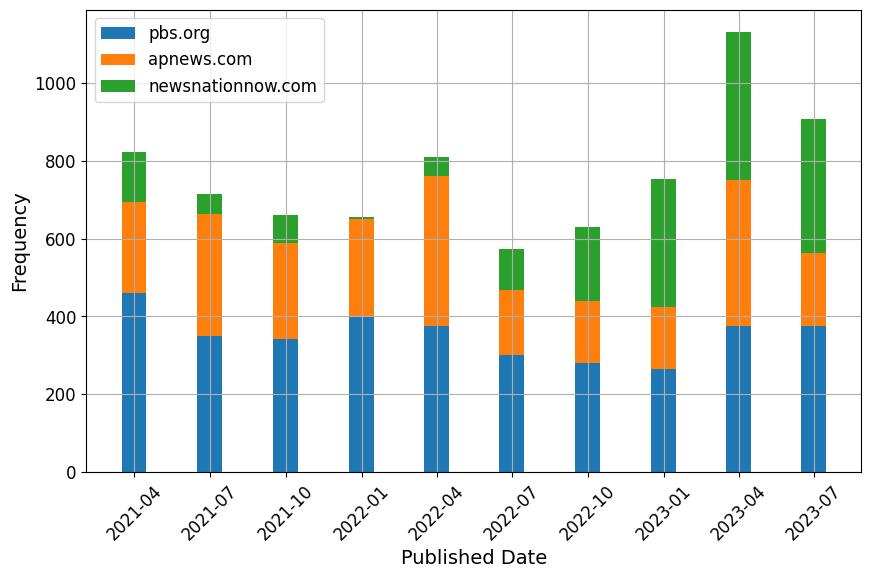}
\end{center}
\vspace*{-.2in}
\caption{Center Outlets Aggregate Distribution}
\label{Fig:center-outlets-distro}
\vspace*{-.3in}
\end{figure}

Figure~\ref{Fig:left-outlets-distro}-~\ref{Fig:center-outlets-distro} presents final state of training data \citep{OKC2022}, demonstrating even distributions across different time periods and outlet grouping.

\vspace*{-.1in}

\vspace*{-.35in}

\vspace*{.2in}

\subsection{Application Data Distribution Figures} 
\label{Appendix:App-Data}
\begin{figure}[H]
\begin{center}
\includegraphics[width=7.6cm,height=5.6cm]{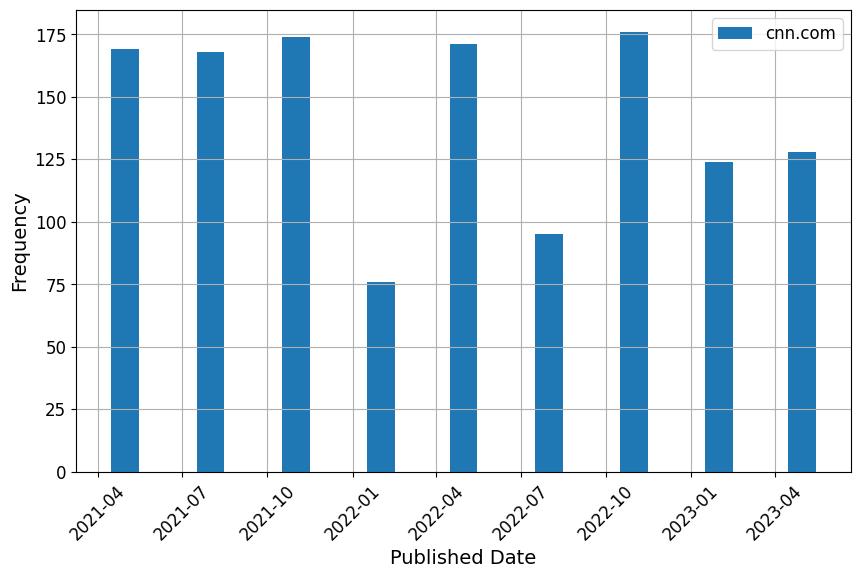}
\end{center}
\vspace*{-.2in}
\caption{CNN Articles: Distribution by Period}
\label{Fig:CNN-distro-by-period}
\end{figure}

\begin{figure}[H]
\begin{center}
\includegraphics[width=7.6cm,height=5.6cm]{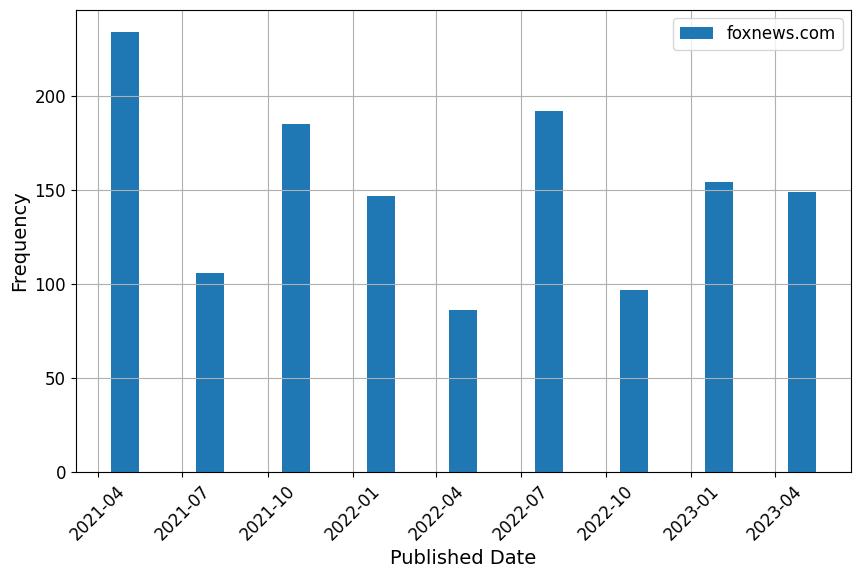}
\end{center}
\vspace*{-.2in}
\caption{FOX Articles: Distribution by Period}
\label{Fig:FOX-distro-by-period}
\end{figure}

The histograms below show the articles queried from FOX and CNN. These articles are used to apply the models developed throughout the study to external news outlets. As can be seen in Figures ~\ref{Fig:CNN-distro-by-period} and ~\ref{Fig:FOX-distro-by-period}, roughly 1500 articles are queried from each outlet to be evenly distributed throughout the 3 year interval explored.

\subsection{Convolutional NN Model Architecture}
\label{Appendix:convolutional-model-architecture}
The code shown in Figure~\ref{fig:CNN-Arch} defines the architecture for the convolutional neural network model, which is largely inspired by the work of \citet{atmosera2023}.
First an Embedding layer converts
integer-encoded vocabulary into dense vectors of fixed size. This layer efficiently 
handles the vast vocabulary of text data, providing
meaningful representations of words 
that capture semantic similarities based on their context within the corpus. This 
dense vector representation allows the model to interpret text input effectively,
facilitating 
identification of patterns relevant to classification tasks.
\begin{figure} [H]
    \centering
    \begin{lstlisting}[]
from tensorflow.keras.models import Sequential

from tensorflow.keras.layers import Embedding, Flatten, Dense, Conv1D, MaxPooling1D, GlobalMaxPooling1D

num_classes = 3  

model = Sequential()

# Embedding layer
model.add(Embedding(100000, 32))

# First convolutional layer
model.add(Conv1D(32, 7, activation='relu'))

# First pooling layer
model.add(MaxPooling1D(5))

# Second convolutional layer
model.add(Conv1D(32, 7, activation='relu'))

# Global max pooling layer
model.add(GlobalMaxPooling1D()) 

model.add(Dense(num_classes, activation='softmax'))  
model.compile(loss= 
    'sparse_categorical_crossentropy',  
              optimizer='adam',
              metrics=['accuracy'])
    \end{lstlisting}
    \caption{CNN Architecture Code}
    \label{fig:CNN-Arch}
\end{figure}

Following the Embedding layer are two sets of one-dimensional convolution layers and Max Pooling layers. The convolution layers apply convolutional operations to the embedded word vectors, using filters to extract local patterns (such as the presence of specific n-grams) 
indicative of the text's class. The rectified linear unit (ReLU) activation function ensures that the model captures nonlinear relationships between these features. Each convolution layer is followed by a Max Pooling layer, which reduces the dimensionality of the data by retaining only the most prominent features, thus improving computational efficiency and helping to prevent overfitting.

After the second convolution and pooling sequence, a Global Max Pooling layer aggregates the most significant features from across the entire text, ensuring that the model's final predictions are informed by the most impactful elements of the input data. The architecture culminates in a Dense layer with a Soft Max activation function, which maps the extracted features to probabilities across the three classes, allowing the model to quantify and output the distinctions noted between classes explored. The model is trained through five separate epochs, making
use of a validation dataset to progressively increase its accuracy.

\subsection{Convolutional NN Model Training}
\label{Appendix:convolutional-model-training}
\begin{figure}[H]
\begin{center}
\includegraphics[width=8cm,height=6.5cm]{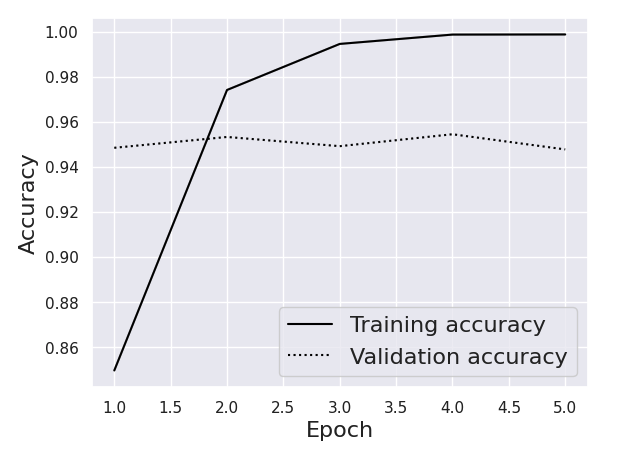}
\end{center}
\vspace*{-.2in}
\caption{Convolutional NN Model Training Process: Training and Validation Accuracy}
\label{Fig:convolutional-model-training}
\end{figure}

The plot in Figure~\ref{Fig:convolutional-model-training} shows the progression of the convolutional NN model’s training and validation accuracy throughout the five training epochs. Training accuracy defines the model's ability to precisely classify articles it recurrently sees throughout each epoch, whereas validation accuracy refers to the model's ability to generalize to unseen data. It is common for validation accuracy and training accuracy to initially increase together. When validation accuracy plateaus, while training accuracy continues to increase, the model begins to overfit to its training data and looses its ability to generalize to external data (e.g. validation data). Five epochs are sufficient to train the model as the validation accuracy begins to plateau around the fourth epoch. 

\end{document}